\documentclass[letterpaper, 10 pt, conference]{ieeeconf}  % Comment this line out if you need a4paper

\IEEEoverridecommandlockouts                              % This command is only needed if 
\title{\LARGE \bf
Where to Fetch: Extracting Visual Scene Representation\\ 
from Large Pre-Trained Models for Robotic Goal Navigation

}
\author{Yu Li, Dayou Li, Chenkun Zhao, Ruifeng Wang, Ran Song and Wei Zhang${^*}$
\thanks{Yu Li, Dayou Li, Chenkun Zhao, Ran Song and Wei Zhang are with the School of Control Science and Engineering, Shandong University, Jinan, China.}
\thanks{Ruifeng Wang is with Campus Security Smart Brain Shandong Engineering Research Center, Jinan, China}% <-this % stops a space
\thanks{*Corresponding author: Wei Zhang (email: davidzhang@sdu.edu.cn)}
}

\usepackage{listings}
\usepackage{lipsum}
\usepackage{tcolorbox}
\usepackage{graphicx}
\usepackage{multicol}
\usepackage{multirow}
\usepackage{subfigure}
\usepackage{tabularx}
\usepackage{colortbl}
\usepackage{hyperref}
\usepackage{booktabs}

\begin{document}

\maketitle
\thispagestyle{empty}
\pagestyle{empty}

%%%%%%%%%%%%%%%%%%%%%%%%%%%%%%%%%%%%%%%%%%%%%%%%%%%%%%%%%%%%%%%%%%%%%%%%%%%%%%%%
\begin{abstract}

To complete a complex task where a robot navigates to a goal object and fetches it, the robot needs to have a good understanding of the instructions and the surrounding environment. 
Large pre-trained models have shown capabilities to interpret tasks defined via language descriptions.
However, previous methods attempting to integrate large pre-trained models with daily tasks are not competent in many robotic goal navigation tasks due to poor understanding of the environment.  
In this work, we present a visual scene representation built with large-scale visual language models to form a feature representation of the environment capable of handling natural language queries.
Combined with large language models, this method can parse language instructions into action sequences for a robot to follow, and accomplish goal navigation with querying the scene representation.
Experiments demonstrate that our method enables the robot to follow a wide range of instructions and complete complex goal navigation tasks. Video of the real-world experiments is at \url{https://youtu.be/Qqo4hox0_vU}.

\end{abstract}

%%%%%%%%%%%%%%%%%%%%%%%%%%%%%%%%%%%%%%%%%%%%%%%%%%%%%%%%%%%%%%%%%%%%%%%%%%%%%%%%
\section{Introduction}
For a robot to locate and fetch something under language command given by humans, it should possess the capacity to comprehend language descriptions of the tasks, and perceive the environment to perform goal navigation to find the target objects. 
% Robots epitomize humanity's relentless pursuit of intelligence. They are anticipated to possess the capacity to comprehend language, perceive environments, and aid humans in accomplishing various tasks. 
% This requires robots to be able to comprehend commands based on natural language and execute them in real environments. 
For instance, if the robot is asked to ``throw the bottle into the dustbin", it must accurately acquire the positions of the relevant objects mentioned in the command, which are ``bottle" and ``dustbin'', and subsequently determine the sequence of actions necessary to carry out this instruction. 
In this case, the robot should first locate and fetch the bottle, then navigate to the dustbin, and finally put the bottle into the dustbin. 
% For instance, for the instruction ``bring the coke can to the counter", the robot must accurately acquire the positions of the relevant objects mentioned in the command, such as "coke" and "counter", and subsequently determine the sequence of actions necessary to carry out the instruction. 

\begin{figure}
    \vspace{0.5em}
    \centering
    \includegraphics[width=1\columnwidth]{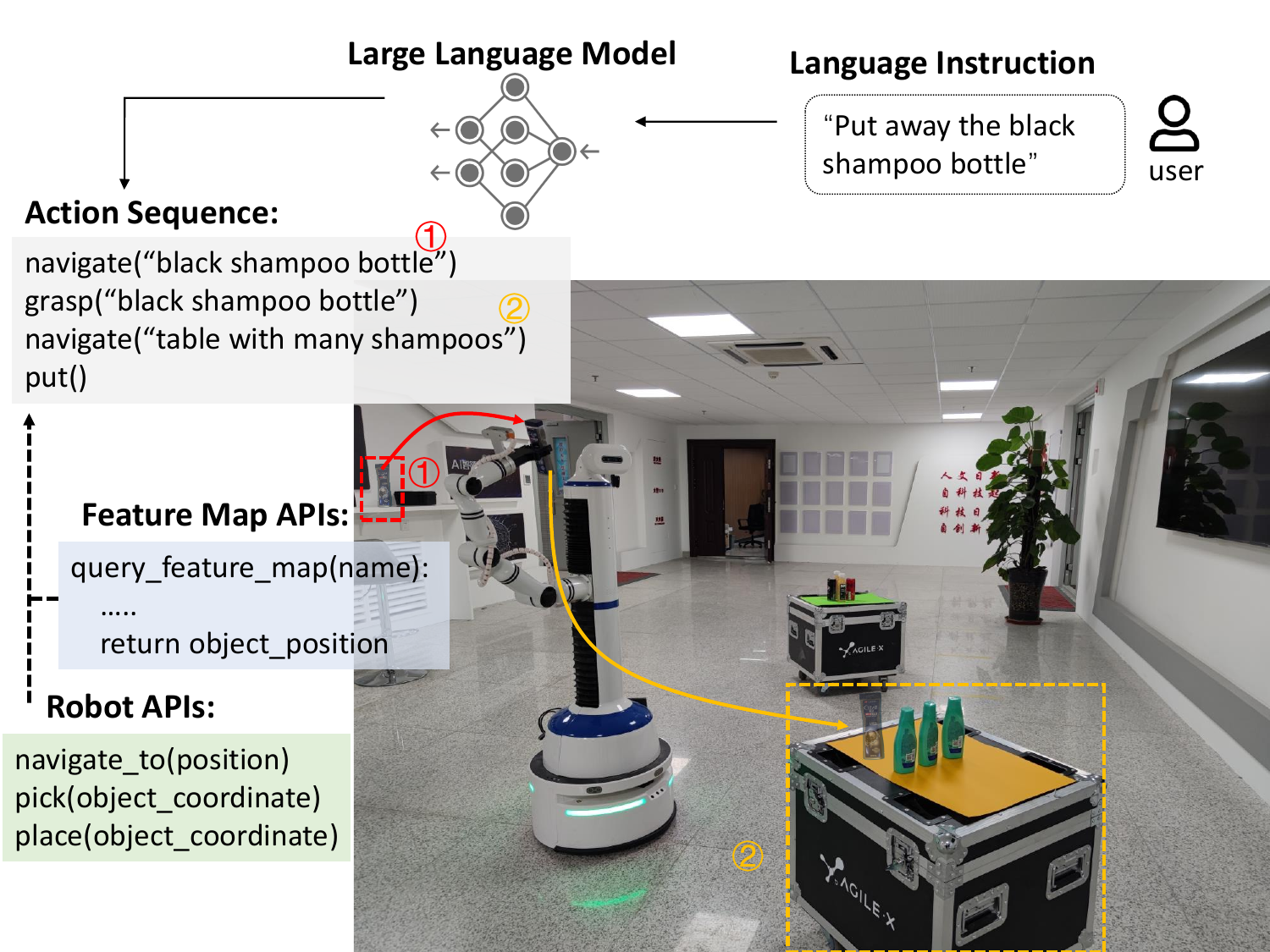}
    \caption{With visual scene representation and LLMs, the robot can locate objects with language descriptions and interpret the language instruction into action sequence to carry out the tasks by calling appropriate APIs.}
    \label{fig:enter-label}
    \vspace{-1em}
    \label{fig:cover}
\end{figure}

Understanding the semantics of a scene is a crucial foundation for goal-oriented navigation tasks under language instructions (see Fig. \ref{fig:cover}). 
Many approaches \cite{mousavian2019visual,du2020learning} have been proposed to accomplish goal navigation to find the target object by searching for it through active explorations. However, such explorations can be repetitive and less efficient. %inefficient. 
% Persistent storage of semantic information via semantic maps is a reasonable choice. 
% Exploiting semantic information via semantic maps is a reasonable choice to avoid repetitive explorations. %though it may require extra efforts to build it.
Integrating semantic simultaneous location and mapping (SLAM) with goal navigation \cite{mccormac2017semanticfusion} is a reasonable choice to avoid repetitive explorations, but it can only work on a close-set class of objects.
In real-world, objects are too diverse for the close-set to cover.

% Semantic simultaneous location and mapping (SLAM) \cite{mccormac2017semanticfusion} based on methods like semantic segmentation can return a map with object locations, but it can only work on a close-set class for objects.

%beneficial to robotic systems.
% Admittedly, multiple approaches of Object Goal Navigation \cite{mousavian2019visual,du2020learning} can avoid map constructions by searching for target points through active explorations, but these explorations could be repetitive and less efficient. %inefficient. 

% 需要写这句话么？
% For example, when we want to search for something whose class are not listed in the dataset, like ``black cubic bottle'', these methods would be unable to search and fetch the goal.%ineffective. 

% However, there cannot be only objects in the environment within the close-set, making this method unable to record all the objects. 
% Furthermore, it cannot handle human queries based on natural language such as ``a black bottle", a class of items described by definitive terms and not inside the class set.
% Furthermore, it cannot handle human queries based on language instructions, as which describe items out of the definitive class set. 
% 第一段加参考文献. 整体的逻辑：批现在方法只能做close-set的object navigation

Large pre-trained models such as large-scale visual language models (VLMs) \cite{pmlr-v139-radford21a, alayrac2022flamingo} and large language models (LLMs) \cite{gpt3, openai2024gpt4} have demonstrated great ability in language comprehension, recognition of common sense, and logical reasoning. 
Furthermore, the generalization of large pre-trained models can enable robots to accept complex instructions \cite{lmnavshah2022}.
For example, many works \cite{NEURIPS2022_d0b8f0c8, gadre2022cow, pmlr-v202-zhou23r} demonstrated that large pre-trained models can be used as a substantial component for open-vocabulary goal navigation. 
The models used in such methods aim to provide a match between language descriptions and raw visual observation of robots. % showing the potential of characterizing the semantics of the environment. 
However, matching descriptions with images is not sufficient to locate and fetch a particular object, as it only leads to where the images are taken rather than the coordinate of the target object.

\begin{figure*}[t]
	\centering
	\includegraphics[width=0.95\textwidth]{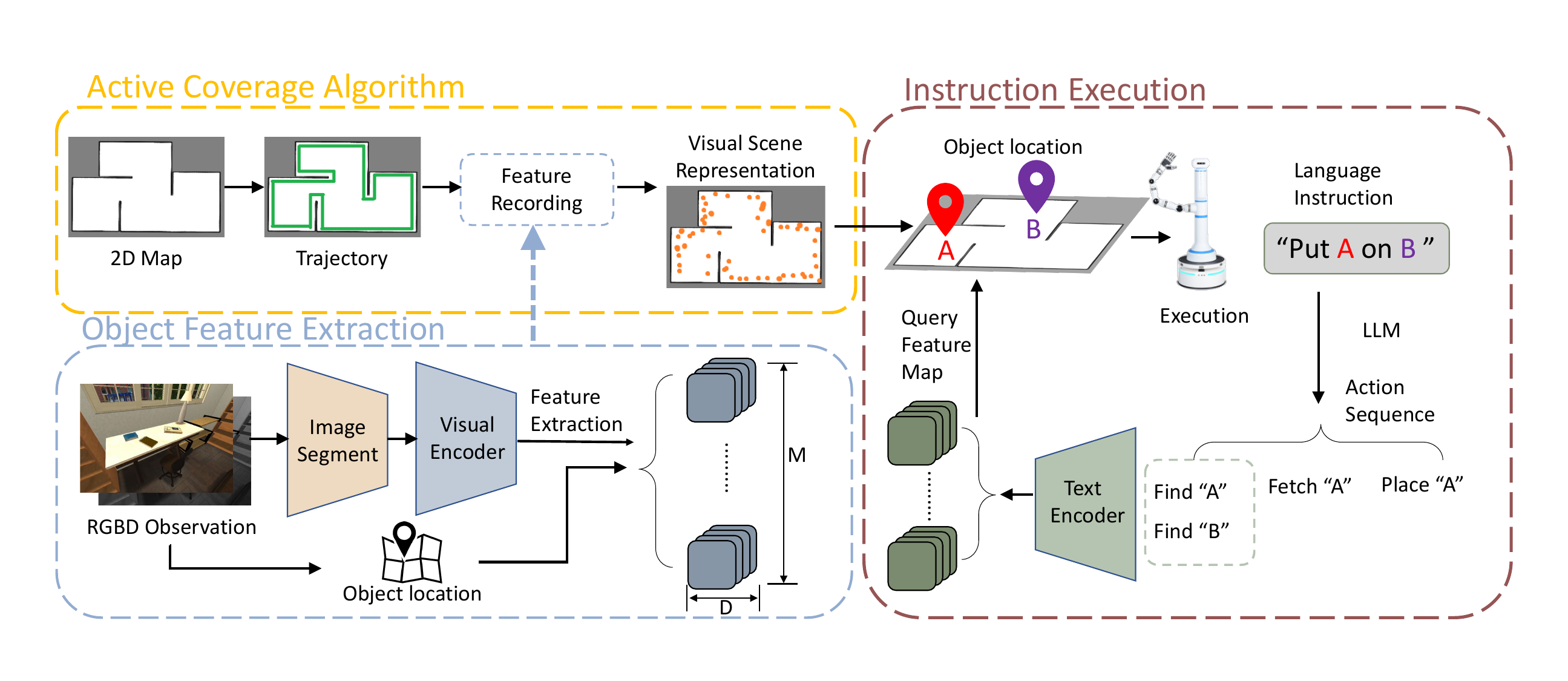}
	\caption{System overview. The visual scene representation is built during the active coverage process as the robot moves to cover the environment. 
It merges the object visual feature and location of objects into the construction.
Once the construction is complete, the robot can use the large language model to break down language instruction into an action sequence and locate the target object with textual description.}

	\label{fig:system_overview}
	\vspace{-1em}
\end{figure*}

This paper presents a robotic goal navigation method based on large pre-trained models, which consists of three parts. Specifically, the first part is object feature extraction from robot observation. 
It takes the RGBD observation as input. By applying image segmentation and the visual encoder of the visual language model, it extracts visual features for almost all objects observed.
At the same time, it calculates the coordinates of the objects and records them with the extracted feature. The second part is an active coverage algorithm, which takes the map of the environment and the output of the object feature extraction as inputs. 
It extracts the path to cover all the objects in the environment and navigates the robot along it. 
During the coverage process, object features are recorded to create a Visual Scene Representation (VSR) of the environment. 

% The second is an active coverage algorithm, which uses the map of the environment and the output of the object feature extraction to cover all the points in the environment necessary to create a complete feature set of the environment. 
% While collecting the features extracted during the active coverage process, a Visual Scene Representation (VSR) is obtained.

The third part is instruction execution, where we use a LLM to analyse the instruction, extract target descriptions, and work with the VSR to perform goal navigation and other tasks with the robotic arm.

% In this paper, we present a visual scene representation (VSR) built with visual language models (VLM). 
% The VSP can be used as a map for object goal navigation with language queries.
% In a word, LLM offers the possibility of language-based interaction for robots, which, aided by prompt engineering, allows robots to extract natural language into action sequences. In this paper, we present a visual-language-feature-map, built with pure off-the-shelf methods SAM and CLIP, to form a feature representation of the environment, capable of handling natural language queries.
% Our method, combined with LLM, can comprehend given language instructions and then integrate VSR with downstream tasks such as grasping and placing down objects with robotic arm. %manipulation
In summary, our main contributions are as follows:
% Combined with LLM, this method will be able to expand the range of language instructions for robots to a great extend. In summary, our main contributions are as follows:
\begin{itemize}
% \item An automatic pipeline for constructing visual scene representation. %that can be used for language querying. 
\item A novel pipeline for constructing object level visual representation for robotic goal navigation. %that can be used for language querying. 
\item An active coverage algorithm for targeted locations in the environment that allows a robot to operate autonomously to accomplish VSR construction.
\item A combination of an LLM and VSR to deliver a robot system that executes language-based instructions.
\end{itemize}

\section{Related Work}
% \note{related work should be written in the format of: xxx et al. propose xxx to, and then ...}
\subsection{Semantic Mapping}
Constructing a semantic representation of the environment can be very useful for goal navigation. %robot planning and 
% Traditional SLAM algorithms are well developed.  
Semantic SLAM using neural networks has recently made certain advancements, including semantic segmentation \cite{mccormac2017semanticfusion,8593691}, object detection \cite{zhang2018semantic,8440105}, etc. 
These methods enrich the map representation by adding semantic features to the geometry features constructed by SLAM algorithms. 
% Furthermore, progress has been made in generating more abstract maps, such as scene graphs \cite{wu2021scenegraphfusion}. 
Furthermore, Wu et al. \cite{wu2021scenegraphfusion} proposes scene graphs to generating more abstract maps, making progress in scene representation.
However, existing methods for semantic representations of the environment are limited to pre-defined categories, limiting the application scenarios. 
%that cannot be queried by natural language
By constructing VSR, our work combines the visual features of objects with their physical coordinates, to form a representation that can be used for open-vocabulary goal navigation.
%natural language-based instruction understanding.

\subsection{Goal Navigation}
Goal Navigation plays an important role in embodied AI. Its task is to find a target object in the environment. 
Some approaches \cite{mousavian2019visual,pmlr-v100-sax20a} exploit only the robot's current observations to navigate, while some \cite{ramrakhya2023pirlnav, du2020learning} improve overall efficiency by incorporating the memories of past observations. 
None of these methods build maps or preserve the feature representation of the environment as a whole. 
% Chaplot et al. \cite{chaplot2020object} proposes SemExp to combine object goal navigation with the construction of semantic maps, improves the interpretability of the method and makes it more possible for real-world deployment. But it is limited to work on a close-set class of objects.
Our work aims to plan as well as manipulate based on obtaining coordinates of objects. 
Once the construction of the scene representation is completed, the subsequent tasks can be performed on this basis without the need to re-explore every time.
% The difference between our work and Object Goal Navigation is that our work aims to plan as well as manipulate based on obtaining coordinates of objects. 
% Once the construction of the feature map is completed, the subsequent tasks can be performed on this basis without the need to re-explore every time.

% \subsection{Large Pre-trained Models for Robotics}
The success of large pre-trained models has effectively led to their integration into robotics extending the capabilities of zero-shot learning to several domains, including robot manipulation \cite{mees2022matters, pmlr-v229-huang23b} and reasoning \cite{zitkovich2023rt, gadre2022cow}. 
% The recent success of large pre-trained models has effectively led to their integration into robotics, which, together with visual language pre-training models, extends the capabilities of zero-shot learning to several domains, including robot manipulation \cite{mees2022matters, pmlr-v229-huang23b}, reasoning\cite{zitkovich2023rt, gadre2022cow}. 
For goal navigation, LM-Nav \cite{lmnavshah2022} uses three pre-trained models, including LLM, to parse textual commands and perform the task of landmark-by-landmark navigation in the real world using a constructed topological map. 
% LLM-Planner \cite{song2023llmplanner}, on the other hand, performs navigation tasks by understanding the environment and interacting with LLM in real-time. 
Song et al. \cite{song2023llmplanner} performs navigation tasks by understanding the environment and interacting with LLM in real-time. 
Huang et al. \cite{huang23vlmaps} uses VLMAPS to achieve finer-grained navigation by combining visual SLAM to build obstacle maps with an open vocabulary.
% VLMAPS \cite{huang23vlmaps} achieves finer-grained navigation by combining visual SLAM to build obstacle maps with an open vocabulary.
In contrast, we use LLM to work with the visual scene representation to enable accurate scene understanding, allowing the robot to manipulate objects.

\section{Method}

To find an item, you need to know where it is. We aim to build VSR that matches item features with their coordinates and can be queried using natural language for subsequent tasks.

\subsection{Visual Representation Construction}

Visual representation aims to obtain the feature of each object in the environment using the robot's vision sensors, and it is first necessary to distinguish between the objects. 
Popular Object detection algorithms such as YOLO \cite{YOLO_Redmon_2016_CVPR} and Fast R-CNN \cite{Girshick_2015_ICCV} have excellent performance and have been deployed in many scenarios.
% Object detection algorithms are common solutions, and algorithms such as YOLO \cite{YOLO_Redmon_2016_CVPR} and Fast R-CNN \cite{Girshick_2015_ICCV} have excellent performance. 
However, these methods can only output the fixed categories present in the dataset and usually require fine-tuning, which makes it difficult to ensure the acquisition of a complete set of objects, and is less capable of natural language querying. 
% Open vocabulary object detection algorithms such as VILD \cite{gu2022openvocabulary}, alleviate the limitations of the fixed recognition categories. 
% However, its generalization performance is limited by the region proposal network. 
% no matter what object detection algorithm is used, there is no guarantee that all objects in the environment can be detected.
% However, no matter what object detection algorithm is used, there is no guarantee that all objects in the environment can be detected.
% Another solution is image segmentation. 

Segment Anything Model (SAM) \cite{kirillov2023segany}, whose pre-trained model has excellent zero-shot performance is suitable for us. 
For SAM, it returns a valid segmentation mask for the given prompt.
Based on this, it is possible to obtain images for almost every objects in the environment by adequately prompting the model with points in the images from the robot observation.
% without the need for fine-tuning.

We propose to use SAM to obtain the segmented object images $I_i\in I_{1...N}$ from the visual observation, \textit{N} represents the number of objects obtained by the segmentation in the observed image. 
For segmentation results, we fill the holes inside the segmented images with pixels from the original image.
This helps to avoid obtaining objects with rugged boundaries that may generate unclear meanings. 
% The result is an image that contains more objects, while still maintaining clear boundaries. 
The image alone cannot establish a corresponding mapping relationship with the object described in natural language, we also need to encode the image to obtain the feature representation to establish the connection between the object represented by the image and the natural language in the feature space. 
For the object images obtained, we then use the CLIP model and apply the visual encoder $E_v$ for every object image to obtain feature representation which can be expressed as: 
\begin{equation}
    \Phi_{i}^{D}=[E_v(I_i)| i\in 1...N],
\end{equation}
where \textit{D} is the feature dimension, usually 512.

We also need the coordinates of the object to complete the subsequent localization process. 
The centre pixel $P_i=(u_i, v_i)$ of the instance is transformed to point $p_i=(x_i, y_i, z_i)$ in the map frame by the depth image, the camera intrinsic matrix, and the robot localization. 
The combination of the obtained data $t_i=(\Phi_{i}^{D},p_i)$ is the object-level feature representation of the observed value of the image.

\subsection{Active Coverage Algorithm}
To improve the overall efficiency, we need to construct visual scene representation autonomously. 
In this process, by adopting the active coverage algorithm, the robot autonomously conducts coverage scanning of objects in the environment, completes feature extraction, and achieves VSR construction. 
Our goal is to traverse all objects in the environment, which naturally makes us think of the full coverage algorithms \cite{10.1007/978-981-15-9460-1_20} used by, for example, sweeping robots. 
But our situation is different. 
The robot does not need to cover all the locations in the environment, but only those where the objects are placed.
% , and naturally there are no objects in the open locations as there are no obstacles to pay attention to. 
At the same time, the camera is pointed to the direction of the obstacles, and images of all obstacle directions are recorded in the process.

To this end, we first process the 2D map to obtain the locations where active coverage should be performed, and complete the active coverage of the target area by abstracting the problem to solving the Hamiltonian path problem, an example of the result is Fig. \ref{fig:active_coverage_algorithm}. This scheme contains the following three specific steps:

\begin{figure}[ht]
  \centering
    \subfigure[Original map]{\includegraphics[width=0.32\linewidth]{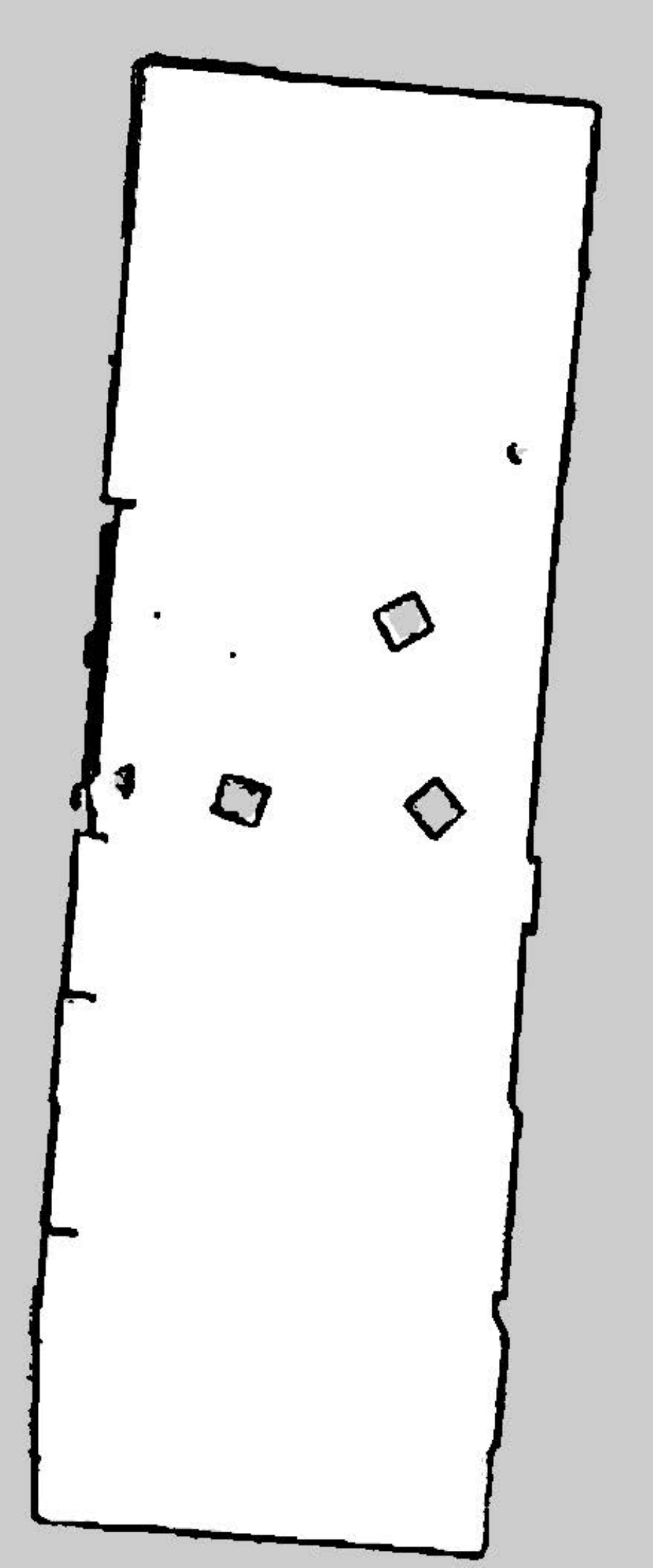}}
    \subfigure[Boundary]{\includegraphics[width=0.32\linewidth]{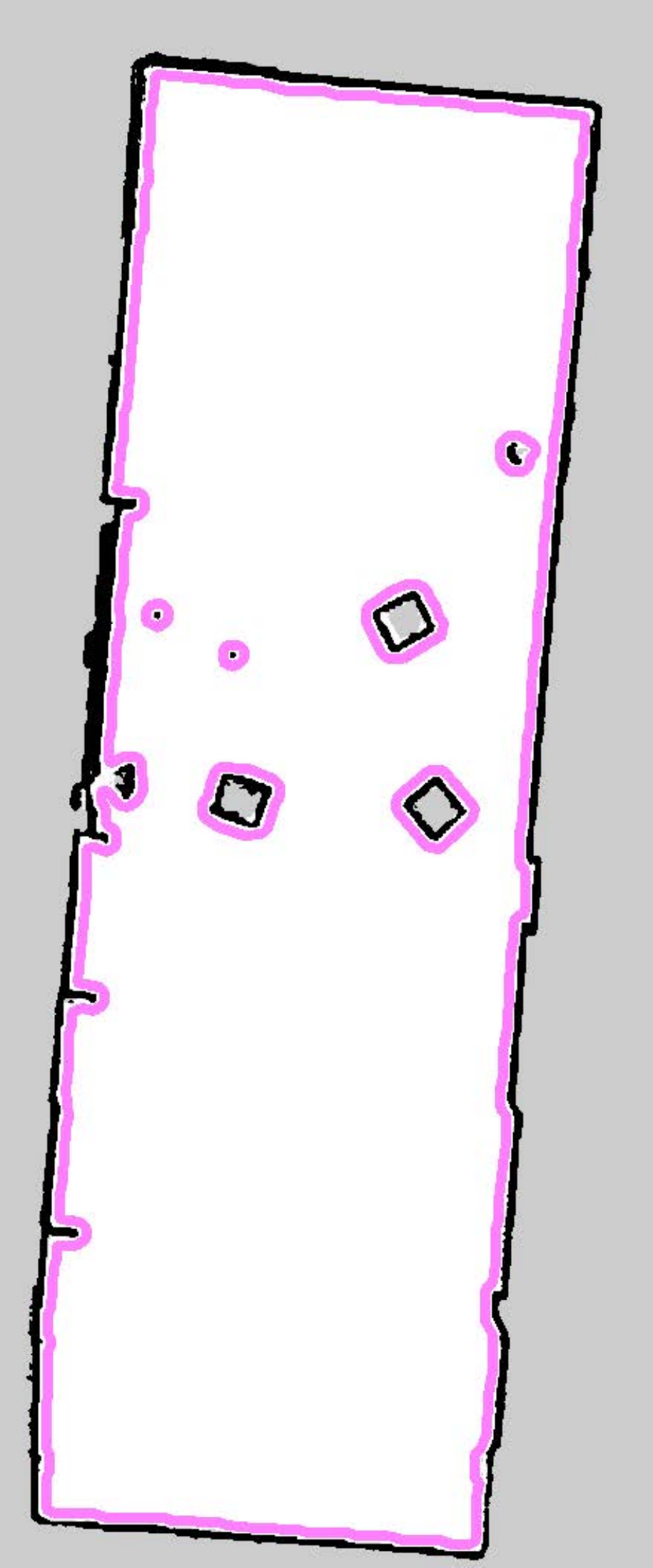}}
    \subfigure[Coverage plan]{\includegraphics[width=0.32\linewidth]{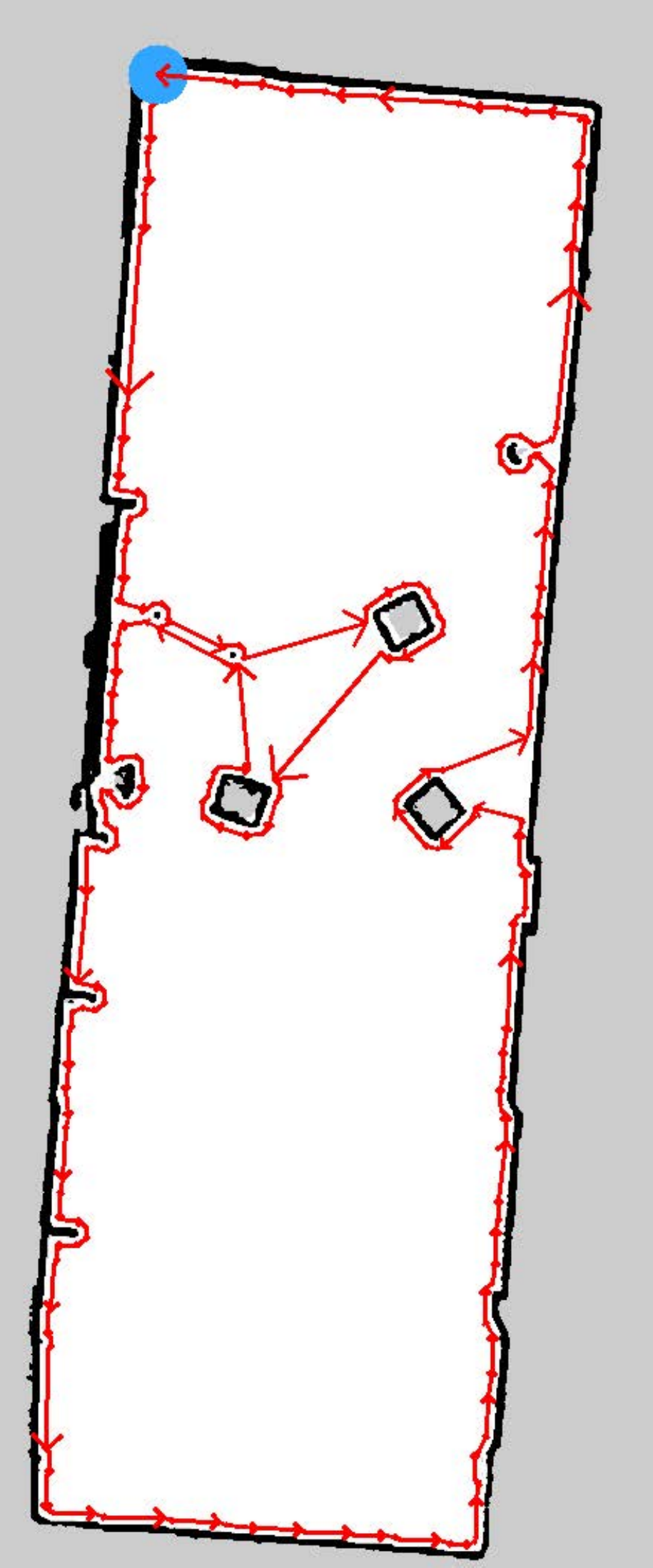}}

  \caption{Active Coverage Algorithm. (a) is the original 2D map. (b) shows result of boundary extraction, with contours drawn in purple. (c) shows coverage path, with blue dot denoting the starting point and the red arrow indicating the trajectory.}
  \label{fig:active_coverage_algorithm}
\end{figure}

\subsubsection{Boundary Extraction}

The 2D map built by the SLAM algorithm indicates the presence or absence of obstacles in the form of grey-scale maps. 
We first binarise the input image, marking obstacles as 1 and non-obstacles as 0. 

Then, by performing the morphological open operation on the binary map image, the noise present in the map as well as some small gaps that are difficult for robots to access can be effectively filtered out. 
Subsequently, the Suzuki algorithm \cite{suzuki1985topological} is run to extract the boundaries. 
It should be noted that in the boundary extraction algorithm, due to its scanning process, the boundary points obtained are naturally arranged in clockwise or counterclockwise order.

% There are still a large number of points in the obtained boundary. If we use such a path for traversal, not only will it require long computation time for path planning, but also the trajectory will also be very tortuous, ending up in low overall operating efficiency. 
To further reduce the amount of points in the obtained boundary, we apply polygon fitting \cite{RAMER1972244} to reduce the number, without impacting on the coverage of the premise. 
% Thus, we apply polygon fitting \cite{RAMER1972244} to reduce the number of points of the contour, without impacting on the coverage of the premise. 
% Through the use of polygons, the trajectory is able to ``cut corners straight", and a more suitable path for the robot to execute is acquired. 

\subsubsection{Use Graph to Connect the Target Path}

After determining the location to be covered, we need to plan a path to complete the traversal in an efficient manner. 
At the same time, it should be noted that when performing the path coverage, the robot will run in the direction of the edge of the target object, rather than facing the target object. 
In order to construct VSR from the images captured by the camera, we need to orient the camera towards the target object, which requires us to coordinate the traversed path with the camera orientation. 
For example, when the camera is facing the right side of the body's forward direction, we should encircle the object in a clockwise direction and vice versa.

As mentioned earlier, the boundary extraction algorithm obtains boundary points in either clockwise or counterclockwise order, so all that needs is to reverse the order of the counterclockwise boundary points to obtain all clockwise boundary points. 
% As for determining the direction of the boundary points, for convex polygons we can take out three consecutive points and determine the order by vector cross multiplication. 
For convex polygons we can take out three consecutive points and determine the order by vector cross multiplication. 
For concave polygons, we can first find the highest point of the polygon, i.e., the point with the largest y-axis coordinate, and then perform the above calculation.

% As mentioned earlier, the boundary extraction algorithm obtains boundary points in either clockwise or counterclockwise order, so all that is needed is to reverse the order of the counterclockwise boundary points to obtain all clockwise boundary points. As for determining the direction of the boundary points, for convex polygons, we can take out three consecutive points and determine the order by vector cross multiplication, but this method cannot be applied to concave polygons.
% For this reason, we adopt the method as shown in Algorithm 3-1, since the vertices of the polygon are already in clockwise or counterclockwise order, we distinguish between clockwise and counterclockwise by searching for the highest point of the polygon, i.e., the point with the largest y axis coordinate, and compare the coordinate-size relationship of the two points on either side of it. Subsequently, after reversing the order of the counterclockwise polygons, the clockwise sorted polygons are obtained.

After sorting, the vertices in the polygon are taken as the nodes, and the adjacency matrix expression of the obtained graph $G=(V,A)$ is constructed as the problem representation by further establishing the connection between the nodes $V = [v_i, i \in {1...M}]$, where the ${\displaystyle M}$ is the number of nodes, and the adjacency matrix ${\displaystyle A}$ is square matrix of order ${\displaystyle M}$, in which $a_{ij}$ represents the cost of the travel from $v_i$ to $v_j$.

Specifically, each node in the polygon around the target object must be connected in clockwise order, and the cost of the edge is set to the Euclidean distance between the two nodes. 
The cost between the rest of the nodes are is to a large integer to indicate that the direction is impassable.
% (set to 0x3F3F3F3F3F as an example of a 64-bit integer). 
Edges between nodes from polygons can be connected bi-directionally, and the cost of these edges is the Euclidean distance between the two points.
% Nodes that are not in the same polygon can be connected bi-directionally to form edges, the cost of all these edges is the Euclidean distance between the two points.

\subsubsection{Problem Description and Solution}

With the graph constructed, the active coverage algorithm can be seen as adding the robot's current position to the graph ${\displaystyle G}$ as the starting point to find a path with a better overall cost in order to cover all the nodes in the graph, and then return to the starting position in order to facilitate the subsequent tasks. 
This problem is abstracted as solving the Hamiltonian path of the graph ${\displaystyle G}$, which is also known as the classical Traveling Salesman Problem (TSP). 
The global path is obtained by solving this problem. %through algorithms such as simulated annealing. 
This path ensures both coverage of the target location and the order in which the target object is encircled.

By recording the object level feature representation $t_i=(\Phi_{i}^{D},p_i)$ during the active coverage process, a collection of these features $C=\{t_i\}$ is obtained as the visual scene representation of the environment.
% The target VSR is obtained by constructing visual representation during active coverage.
% Further, the coordinates under the actual map coordinate system are calculated by passing the path obtained in the map image through parameters including rotation angle, position offset, and map resolution.

\subsection{Combining LLM and VSR}

To perform goal navigation with VSR, we need to query the coordinates of the target objects with their language descriptions.
The natural language description of the object is encoded by the visual language model text encoder $E_l$ to obtain the textual feature description $\Psi_{i}^{D}$, where ${\displaystyle D}$ is the feature dimension and is usually 512, the same as the visual feature.

We calculate the dot product of obtained textual features by the feature set in the feature map, and the inner product $s_i$ is the similarity of each object in the feature map:
\begin{equation}
    s_i = <\psi_{i}^{D}, \Phi_{i}^{D}>.    
\end{equation}
The object with the highest similarity is the target object. 
From the VSR, the coordinates under the map corresponding to the target object can be obtained.
It is important to note that the object query and positioning method relies entirely on pre-trained models to complete the task. 
This approach utilizes the models' zero-shot performance, effectively avoiding the need for label annotation or model training that other methods may require.

The tasks performed by the robot can be considered as a combination of several simple, basic atomic tasks. 
To leverage the power of LLM, we build the application programming interfaces (APIs) for atomic tasks to enable robot manipulation by LLM. 
The atomic tasks are listed in Table \ref{APIs}. 

\begin{table}[]
\renewcommand\arraystretch{2}

\caption{Atomic Task Description}
\label{APIs}

% \resizebox{\linewidth}{!}{ 
\begin{tabular}{|p{0.25\columnwidth}|p{0.65\columnwidth}|}
\hline

Task    \centering            & Description       \\ \hline
Search Object \centering      & Take the object description text as input, use feature map to output the coordinate of the target.  \\ \hline
Navigation   \centering       & Takes the target coordinate as input, navigate autonomously towards to the object.  \\ \hline
Grasp Object \centering       & Accept the textual description of the target object, use the robotic arm to fetch the object with the help of RGBD camera. \\ \hline
Placement   \centering        & Accept the textual description of the target location, manipulate the robot arm to place the fetched object at the target location.    \\ \hline
Other basic tasks  \centering & Including high-level APIs related to robot movement and manipulation of the robot arm.       \\ \hline

\end{tabular}
% }
% \vspace{-\baselineskip}
\end{table}

After decomposing a task, such as ``placing an object in a specific place", it can be effectively decomposed into a combination of simple tasks. 
These, along with the basic application interfaces (APIs) for robot movement and robotic arm manipulation, provide an upper-level interface for robot control. 
This facilitates the use of the LLM for code generation or calling the appropriate API via text generation.

We use prompt engineering for LLM to decompose the task descriptions entered by humans in natural language into atomic task sequences. 
% LLMs have outstanding performance in both textual comprehension and generation, as well as computer language programming. 
% Additionally, LLMs possess a strong understanding of common sense in relation to textual task descriptions, allowing them to respond in a logical and intuitive manner. 
% Additionally,, LLMs poses a strong common sense understanding for the textual descriptions of the tasks we input, and respond to them in a common sense way. 
% Taking the simple task of ``put something somewhere'' as an example, LLM can understand the sentence components and decompose the task into ``get something'', ``go somewhere'' and ``put something''. 
% For models trained with codes from open source repositories, it is possible to write the codes required for the task based on the task input and the provided APIs.
% and then have the robot execute the task.
By decent prompts, the ability of the LLM to handle complex scenarios, especially when mathematical or logical reasoning is required, can be improved. 
For our task it is not enough to do just few-shot prompting, we need to provide reasoning processes in a manner pointed out in Chain of Thought (CoF) \cite{wei2022chain} to guide the language model through the process of reasoning with our input to get the expected output. 
% With the ability of the LLM, we can guide the model to meet the expected output through several samples of prompts, to come to a complete procedures for the task. 
The LLM can be used to guide the model towards the desired output by providing multiple examples as prompts. 
This allows for the development of a complete set of procedures for the task. 
A sample of prompt engineering is provided as followed:

\begin{lstlisting}
&\textbf{User}& &\textbf{Input}&:
Suppose you are a robot that your actions are limited to picking up items with pick(object), placing down items with place(object) and move to object objects or locations with navigate(object). 
Task: Put the apple on the wooden desk.
Explanation: The task could be done by first finding the apple, then moving to the table, finally putting down the apple.
\end{lstlisting}
\clearpage
\begin{lstlisting}
Plan: 1. navigate(``apple"), 2. pick(``apple"), 
3. navigate(``wooden desk"), 4. place(``apple"), 
5. done(). 
Task: Put away the black coke can.
Plan:

&\textbf{Responds}&:
navigate(``black coke can")
pick(``black coke can")
navigate(``appropriate storage location")
place(``black coke can")
done()
\end{lstlisting}

\section{Experiments}

\begin{figure*}[t]
	\centering
	\includegraphics[width=0.95\textwidth]{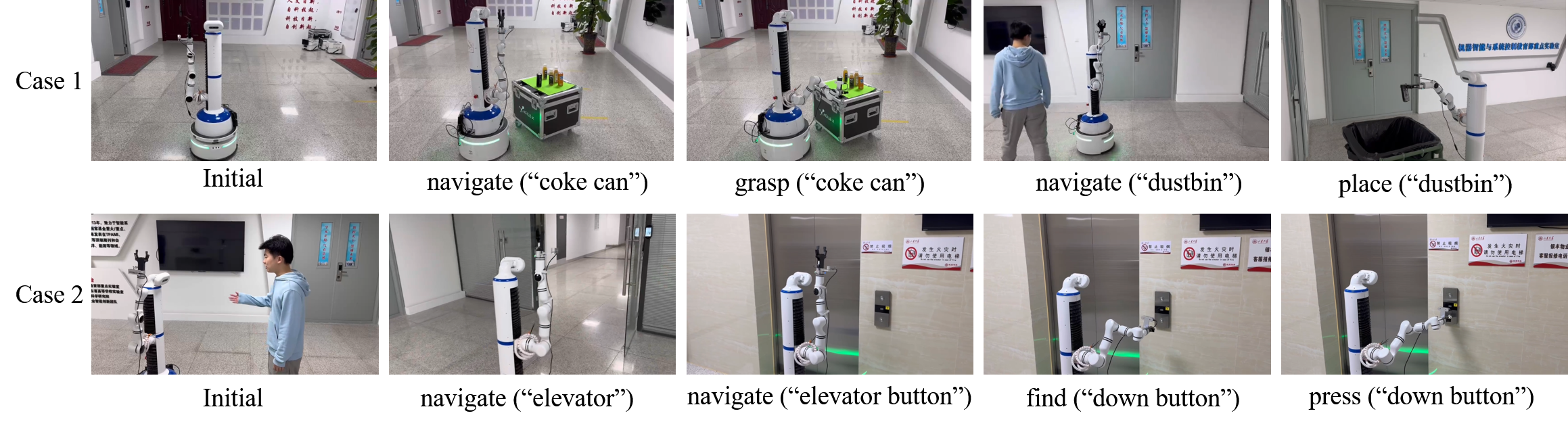}
	\caption{Sample of executing action sequence. 
 For case 1, the robot is asked to ``throw the coke can into the dustbin''. 
 For case 2, the robot is aksed ``I want to go downstairs, can you help''. 
 With pre-trained models, the robot is able to comprehend complex instructions.}

	\label{fig:sample_of_seqence}
	\vspace{-1em}
\end{figure*}

In this section, we evaluate VSR as well as the overall robot system for handling real-world language instructions. %of goal navigation and subsequent tasks with robotic arm.
The experiments were conducted in a real laboratory, and several chests were arbitrarily placed to enrich the scenario.
Video demonstration of the real-world tests can be found at: \url{https://youtu.be/S_QKZqkh0w8}.

% We evaluate our method in real world with a robot following language instructions given by human users.
The robot platform is RM65 with a lightweight 6 degree-of-freedom (DOF) robotic arm mounted on a two-wheel differential mobile platform. 
The mobile platform is equipped with a 2D lidar and an RGBD camera, which are used to avoid obstacles at different heights. 
Two RGBD cameras returning $640 \times 480$ RGBD images are used as main sensors for manipulation tasks with a robotic arm, one is placed at the head of the robot, the other at the tip of the arm.
% The main computing unit is NVIDIA® Jetson AGX Orin™ for on-device AI inference.
The primary computing unit for on-device neural network inference is NVIDIA Jetson AGX Orin. To set up the experiment, we first construct the VSR using the active coverage algorithm.
% The experiment environment setup is an indoor scene with multiple objects at various locations. 

\subsection{Visual Scene Representation}

Performing goal navigation to find the correct object from the given textual query is fundamental to our work. 
The evaluation images are captured by the camera on the head of the robot during the active coverage process, running at 30FPS. 

The effectiveness of the visual feature map is evaluated by determining if the method can accurately return the corresponding image for the given textual input.

% 这写的是啥话
% Finding the right position from the feature map means finding the right observations. We must first see something before knowing where it is.

% We collect an image as observation and use that for feature map construction every 2 seconds. 
% Some images may be blurry due to the acceleration of the robot, but the overall performance is not affected.
% We check the precision by seeing if the method could return the appropriate image for the textual query. 
For query texts, we use the words that describe the elements in the surroundings.
Specifically, We categorize queries into two types, the object finding and location finding. Object finding refers to the ability for finding the correct object, typically something that can be manipulated by the robotic arm. 
Location finding, in this case, demonstrates the ability to locate the correct storage position for objects, like a table or a shelf. 

The effectiveness of the VSR is tested by comparing its querying precision against four other methods capable of zero-shot language-image matching. 
% This involves returning the appropriate image observation from the given textual queries.

\begin{figure*}
    \vspace{0.5em}
    \centering
    \includegraphics[width=0.8\textwidth]{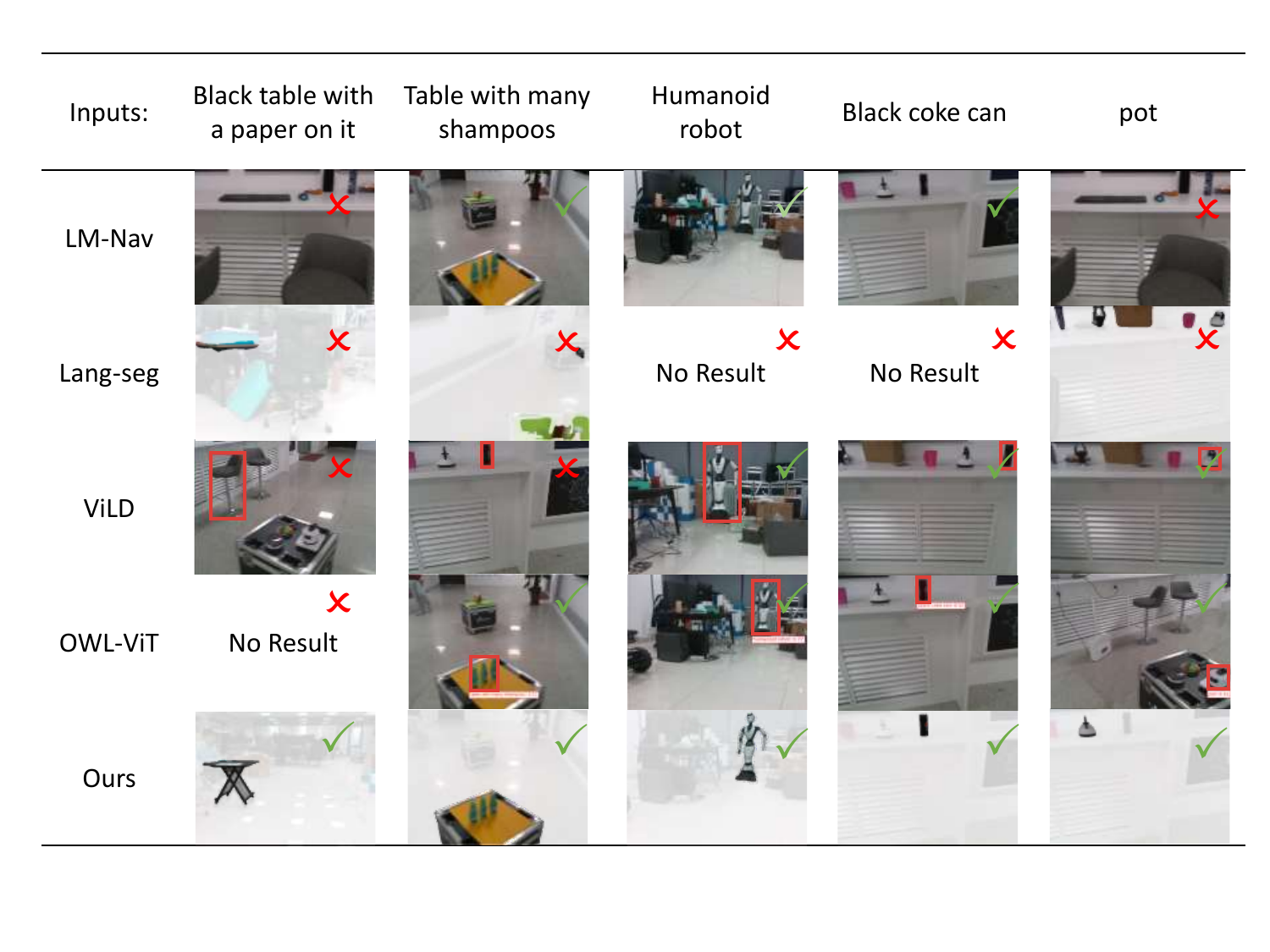}
    \caption{Comparison of different zero-shot language-image matching methods. The results are indicated by a red cross for incorrect matches and a green check mark for correct matches.
        The segment-based methods are presented in red bounding boxes, while the detection-based methods are shown with a transparent background in contrast to the segmented results.
        }
    \label{fig:enter-label}
    \vspace{-1em}
    \label{fig:contrast_exp}
\end{figure*}

\begin{itemize}
\item LM-Nav \cite{lmnavshah2022} navigates through a list of landmarks by applying CLIP to find the corresponding image observation with textual descriptions. 
This method also serves as an ablation study for our object finding where we use raw images as input instead of doing feature extraction. 
% parses language instructions into
% creates a graph where image observations of an environment are stored as nodes while the proximity between images are represented as edges. 
% By combining GPT-3 and CLIP, it  and plans on the graph towards corresponding nodes.
\item LSeg \cite{li2022languagedriven} achieves remarkable zero-shot language-driven semantic image segmentation by using a text encoder for descriptive input labels and an image encoder to compute per-pixel embeddings. 
It offers the ability to build scene representations for language queries.

\item VILD \cite{gu2022openvocabulary} implements open-vocabulary object detection by combining region proposal network (RPN) with open-vocabulary image classification to form a two-stage object detector. 
 % distills embeddings obtained by applying CLIP or ALIGN to cropped image regions from a class-agnostic region proposal network (RPN)
It is capable of matching objects with novel class description unseen in object detection datasets.

\item OWL-ViT v2 \cite{NEURIPS2023_e6d58fc6} is another outstanding open-vocabulary object detection method. It utilizes Vision Transformers (ViT) as an image encoder, avoiding the drawbacks of PRN.
\end{itemize}

% 如此衡量是否合理？我们的方法应当再调试一下。
\begin{table}[t]
  \setlength\tabcolsep{5.2pt}
    \caption{Comparison with other open-vocabulary baselines on visual-language matching}
  \centering
  \begin{tabular}{lccc}
  \toprule
\multirow{2}[1]{*}{Method}  & \multicolumn{2}{c}{Target} &          \multirow{2}[1]{*}{Success Rate (\%)}             \\%& Task \\                
                    \cmidrule(lr){2-3}
                        &   Object                 & Location             \\%& Success Rate\\
\midrule
LM-Nav \cite{lmnavshah2022}                 & 10                  & 4                & 47                   \\%&   1                 \\
LSeg \cite{li2022languagedriven}                    & 4                  & 3               & 23                      \\%&   1             \\ % docker解决
VILD \cite{gu2022openvocabulary}                    & 13                  & 4                & 57                   \\%& 1  \\ % colab
OWL-ViT v2 \cite{NEURIPS2023_e6d58fc6}                & 15                  & 6                & 70                   \\%& 1  \\ % colab
\textbf{Ours}           & \textbf{17}         & \textbf{8}      & \textbf{83}        \\%& \textbf{70} \\
% \midrule
% Total Tasks             & 20                  & 10               & -                 \\%& - \\
  \bottomrule
  \end{tabular}

  \label{table:feature_map_exp}
  \vspace{-1em}
\end{table}

We have 20 tasks for objects and 10 for locations.
As the results in Table \ref{table:feature_map_exp} suggest, our method has a better performance over all baselines, with a success rate of 85\% for object query, and 80\% for position query. 

We list some of the matching result in Fig \ref{fig:contrast_exp}.
LM-Nav performs poorly because the raw image is not suitable for textual description of objects that only occupy a small part of the image. 
Furthermore, this method could only navigate to locations where the images are taken, which can not help with downstream tasks such as grasping with the robotic arm.
For the segment-based method LSeg, the method is trained on more limited datasets and could only perform well with texts similar to the given categories from the datasets.
Open-vocabulary object detection methods have fair performances but contain more false predictions and are less able to query locations where the target image should be many objects stacked together.

% So lm-nav maybe can find the object in the image, but cannot perform manipulation with object-because do not know the precise position?!

\subsection{Instruction Comprehension}

% VSR enables robot to do goal navigation
With the onboard robotic arm, the robot is possible to do more tasks besides goal navigation.
Once the feature representation of the environment is collected, we pass the control to LLM, or more specifically GPT-4 \cite{openai2024gpt4}, to follow the language instructions.

The robot in this scenario is supposed to follow a variety of instructions. 
From simple instructions such as finding something, to more complex ones aimed at helping humans with real-world tasks like cleaning up the table.

Take the latter as an example, the robot in this scenario is supposed to tidy up the table, by moving the objects on the table to the place for proper storage. 
For instance, a coke can on the table should be moved to the place with many other drinks. 
To meet this end, the robot is given simple instructions such as ``put away the black Coke can''. 
By few-shot prompting, robot is able to break down the instructions into a sequence of tasks. 
The ``black coke can" is then successfully found and then placed to the proper storage place described as ``table with many coke cans". 
The same goes for the shampoo bottle and the pot.

% As long as the object query procedure works, the tasks the 

Overall, the above results demonstrate the effectiveness of visual feature maps and our pipeline for executing language instructions in real world environment. 
Moreover, it shows the ability to handle arbitrary tasks given by humans, indicating the potential for a more general AI embodiment.

\section{Conclusion}

In this work, we present a goal navigation method based on large pre-trained models. %visual scene representation that is capable of handling language queries. 
It helps to build a robotic system that finds the target objects and executes language instructions given by humans, showing great potential for assisting humans in daily tasks.
% Combined with large language model, it helps to build a robotic system that executes language instructions given by humans, showing great potential for assisting humans in daily tasks.

For future work, we will address the challenge of handling dynamic scenarios where objects may not be in their original positions when the map was built. 
We will also look into building a more interactive system to manage even more complex tasks.


%%%%%%%%%%%%%%%%%%%%%%%%%%%%%%%%%%%%%%%%%%%%%%%%%%%%%%%%%%%%%%%%%%%%%%%%%%%%%%%%

%%%%%%%%%%%%%%%%%%%%%%%%%%%%%%%%%%%%%%%%%%%%%%%%%%%%%%%%%%%%%%%%%%%%%%%%%%%%%%%%



% \section*{ACKNOWLEDGMENT}

% The preferred spelling of the word ÒacknowledgmentÓ in America is without an ÒeÓ after the ÒgÓ. Avoid the stilted expression, ÒOne of us (R. B. G.) thanks . . .Ó  Instead, try ÒR. B. G. thanksÓ. Put sponsor acknowledgments in the unnumbered footnote on the first page.

%%%%%%%%%%%%%%%%%%%%%%%%%%%%%%%%%%%%%%%%%%%%%%%%%%%%%%%%%%%%%%%%%%%%%%%%%%%%%%%%

% References are important to the reader; therefore, each citation must be complete and correct. If at all possible, references should be commonly available publications.

{\small
\bibliographystyle{ieeetr}
\bibliography{my_bib}
}

\end{document}